\documentclass[review]{elsarticle}
\usepackage{eqlist}
\usepackage{lineno,hyperref}
\usepackage{chngpage}
\usepackage{multirow}
\usepackage{tabularx}
\usepackage{booktabs}
\usepackage{array}
\usepackage{amsmath}
\usepackage{amsthm} 
\usepackage[refpage]{nomencl}
\makenomenclature
\newtheorem{definition}{Definition}[section]
\newtheorem{lemma}{Lemma}[section] 

\newtheorem*{myproof}{Proof}
\modulolinenumbers[5]
\biboptions{numbers,sort&compress}

\begin{document}

\begin{frontmatter}

\title{A correlation coefficient of belief functions}

\author[address1]{Wen Jiang\corref{label1}}

\address[address1]{School of Electronics and Information, Northwestern Polytechnical University, Xi'an, Shannxi, 710072, China}
\cortext[label1]{Corresponding author: School of Electronics and Information, Northwestern Polytechnical University, Xi'an, Shannxi, 710072, China. Tel:+86 029 88431267; fax:+86 029 88431267. E-mail address: jiangwen@nwpu.edu.cn; jiangwenpaper@hotmail.com}

\begin{abstract}
How to manage conflict is still an open issue in Dempster-Shafer evidence theory. The correlation coefficient can be used to measure the similarity of evidence in Dempster-Shafer evidence theory. However, existing correlation coefficients of belief functions have some shortcomings. In this paper, a new correlation coefficient is proposed with many desirable properties. One of its applications is to measure the conflict degree among belief functions. Some numerical examples and comparisons demonstrate the effectiveness of the correlation coefficient.

\end{abstract}


\end{frontmatter}

\section{Introduction}
Dempster-Shafer evidence theory (D-S theory) \cite{Dempster1966Upper,shafer1976mathematical} is widely used in many real applications\cite{Yager2016Evaluating,Majb2016AnEvidential,Islam2017Water,Moenks2016Information,Perez2016Using,Denoeux2016Evidential} due to its advantages in handling uncertain information, since decision-relevant information is often uncertain in real systems\cite{Pedrycz2016Data,Pedrycz2016Multiobjective,Frikha2014On}. However, in D-S theory, the results with Dempster's combination rule are counterintuitive\cite{Zadeh1986A} when the given evidence highly conflict with each other. Until now, how to manage conflict is an open issue in D-S theory. In recent years, hundreds of methods have been proposed to address this issue\cite{Chin2015Weighted,Yang2013Evidential,DengXY2016IEEE,Yang2016Anew,Jiang2016Conflicting,Wangjw2016evidence,Moenks2016Information,Zhao2016anovel,Lin2016Anew,Oliveira2016Amulticriteria,Lefevre2013How}. These solutions are generally divided into two categories: one is to modify the combination rule and redistribute the conflict; the other is to modify the data before combination and keep the combination rule unchanged.

Obviously, how to measure the degree of conflict between the evidences is the first step, since we need to know whether the evidence to be combined is in conflict before doing anything in conflict management\cite{Destercke2013Toward,Sarabi2013Information,Yu2015Animproved}. So far there are no general mechanisms to measure the degree of conflict other than the classical conflict coefficient $k$. But since the classical conflict coefficient $k$ is the mass of the combined belief assigned to the emptyset and ignores the difference between the focal elements, using $k$ to indicate a conflict between the evidences may be incorrect. Several conflict measures, such as Jousselme's evidence distance\cite{jousselme2001new}, Liu's two-dimensional conflict model\cite{Liu2006Analyzing}, Song \emph{et al.}'s conflict measurement based on correlation coefficient\cite{song2014coefficient}, have been proposed to measure the conflict in D-S theory. Although some improvements have been made, there are still some shortcomings in the existing conflict measure methods. How to measure the the degree of conflict between the evidences is not yet solved. In D-S theory, the conflict simultaneously contains the non-intersection and the difference among the focal elements\cite{Liu2006Analyzing}. Only when these two factors are considered simultaneously, the effective measure of conflict can be realized. In this paper, we try to measure the conflict from the perspective of the relevance of the evidence, based on simultaneously considering the non-intersection and the difference among the focal elements.

A correlation coefficient can quantify some types of correlation relationship between two or more random variables or observed data values. In D-S theory, a correlation coefficient is usually used to measure the similarity or relevance of evidence, which can be applied in conflict management, evidence reliability analysis, classification, etc\cite{Ma2015Combination,song2014coefficient}. Recently, various types of correlation coefficient are presented. For example, in \cite{Zhang2016Stumble}, a correlation coefficient was introduced to calculate the similarity of template data and detected data, then the basic probability assignments (BPAs) were obtained based on classification results for fault diagnosis. In \cite{Ma2015Combination}, a correlation coefficient is proposed based on the fuzzy nearness to characterize the divergence degree between two basic probability assignments (BPAs). In \cite{song2014coefficient}, Song \emph{et al}. defined a correlation coefficient to measure the conflict degree of evidences. Moreover, some different correlation coefficients are proposed respectively according to specific applications \cite{Zhu2012Information,Zhang2012Improvement,Zhang2016Stumble}. In this paper, a new correlation coefficient, which takes into consider both the non-intersection and the difference among the focal elements, is proposed. One of its applications is to measure the conflict degree among belief functions. Based on this, a new conflict coefficient is defined. Some numerical examples illustrate that the proposed correlation coefficient could effectively measure the conflict degree among belief functions.

The paper is organised as follows. In section 2, the preliminaries D-S theory and the existing conflicting measurement are briefly introduced. Section 3 presents the new correlation coefficient and proofs many desirable properties. In section 4, some numerical examples are illustrated to show the efficiency of the proposed coefficient. Finally, a brief conclusion is made in Section 5.

\section{Preliminaries}

In this section, some preliminaries are briefly introduced.

\subsection{Dempster-Shafer evidence theory}
D-S theory was introduced by Dempster \cite{Dempster1966Upper}, then developed by Shafer \cite{shafer1976mathematical}. Owing to its outstanding performance in uncertainty model and process, this theory is widely used in many fields \cite{Zhang2016ANP,Jiang2016sensor,mo2016generalized}.


\begin{definition}
Let $\Theta = \{ {\theta _1},{\theta _2}, \cdots {\theta _i}, \cdots ,{\theta _N}\}$ be a finite nonempty set of mutually exclusive hypothesises, called a discernment frame. The power set of $\Theta$ , ${2^{\Theta}}$, is indicated as:
\begin{equation}\label{2^X }
{2^{\Theta}} = \{ \emptyset,\{ {\theta _1}\} , \cdots \{ {\theta _N}\} ,\{ {\theta _1},{\theta _2}\} , \cdots ,\{ {\theta _1},{\theta _2}, \cdots {\theta _i}\} , \cdots ,\Theta\}
\end{equation}
\end{definition}

\begin{definition}
A mass function is a mapping $\mathbf{m}$ from ${2^{\Theta}}$ to [0,1], formally noted by:
\begin{equation}\label{mass}
{\bf{m}}:{2^{\Theta}}   \to [0,1]
\end{equation}
which satisfies the following condition:
\begin{equation}\label{M}
\begin{array}{*{20}c}
   {m(\emptyset ) = 0} & {and} & {\sum\limits_{A \in {2^{\Theta}}  } {m(A)}  = 1}  \\
\end{array}
\end{equation}
When m(A)$>$0, A is called a focal element of the mass function.
\end{definition}

In D-S theory, a mass function is also called a basic probability assignment (BPA). Given a piece of evidence with a belief between [0,1], noted by $m(\cdot)$, is assigned to the subset of $\Theta$. The value of 0 means no belief in a hypothesis, while the value of 1 means a total belief. And a value between [0,1] indicates partial belief.
\begin{definition}
Evidence combination in D-S theory is noted as $\oplus$. Assume that there are two BPAs indicated by $m_1$ and $m_2$, the evidence combination of the two BPAs with Dempster's combination rule \cite{Dempster1966Upper} is formulated as follows:

\begin{equation} \label{combination rule}
m(A) = \left\{ {\begin{array}{*{20}{c}}
{0\begin{array}{*{20}{c}}
,&{}
\end{array}\begin{array}{*{20}{c}}
{}&{}&{\begin{array}{*{20}{c}}
{}&{}&{}
\end{array}}
\end{array}}\\
{\frac{1}{{1 - k}}\sum\limits_{B \cap C = A} {{m_1}(B){m_2}(C)} }
\end{array}\begin{array}{*{20}{c}}
{}\\
{}
\end{array}} \right.\begin{array}{*{20}{c}}
{}\\
{}
\end{array}\begin{array}{*{20}{c}}
{A = {\emptyset} }\\
{A \ne {\emptyset} }
\end{array}
\end{equation}
with
\begin{equation}\label{dempster k}
k = \sum\limits_{B \cap C = \emptyset } {{m_1}(B){m_2}(C)}
\end{equation}
Where $k$ is a normalization constant, called conflict coefficient because it measures the degree of conflict between $m_1$ and $m_2$.
\end{definition}
$k=0$ corresponds to the absence of conflict between $m_1$ and $m_2$, whereas $k=1$ implies complete contradiction between $m_1$ and $m_2$. Note that the Dempster's rule of combination is only applicable to such two BPAs which satisfy the condition $k<1$.

\subsection{Evidence distance}
Jousselme et al.\cite{jousselme2001new} proposed a distance measure for evidence.

\begin{definition}
Let $m_1$ and $m_2$ be two BPAs on the same frame of discernment $\Theta$, containing N mutually exclusive and exhaustive hypotheses. The distance between $m_1$ and $m_2$ is represented by:
\begin{equation}\label{dij}
{d_{BBA}}({m_1},{m_2}) = \sqrt {\frac{1}{2}{{({{\overrightarrow m }_1} - {{\overrightarrow m }_2})}^T}\mathop D\limits_ =  ({{\overrightarrow m }_1} - {{\overrightarrow m }_2})}
\end{equation}
Where ${{{\overrightarrow m }_1}}$ and ${{{\overrightarrow m }_2}}$ are the respective BPAs in vector notation, and ${\mathop D\limits_ =  }$ is an ${2^N} \times {2^N}$ matrix whose elements are
$D(A,B) = \frac{{\left| {A \cap B} \right|}}{{\left| {A \cup B} \right|}}$, where $A,B \in 2^\Theta$ are derived from $m_1$ and $m_2$, respectively.

\end{definition}

\subsection{Pignistic probability distance}

In the transferable belief model(TBM) \cite{Smets1999TBM}, pignistic probabilities are typically used to make decisions and pignistic probability distance can be used to measure the difference between two bodies of evidence.

\begin{definition}
Let $m$ be a BPA on the frame of discernment $\Theta$. Its associated pignistic probability transformation (PPT) $BetP_m$ is defined as
\begin{equation}\label{ppt}
BetP_m (\omega ) = \sum\limits_{A \in {2^{\Theta}}  ,\omega  \in A} {\frac{1}{{\left| A \right|}}} \frac{{m(A)}}{{1 - m(\emptyset )}}
\end{equation}
where $\left| A \right|$ is the cardinality of subset $A$.
\end{definition}

The PPT process transforms basic probability assignments to probability distributions. Therefore, the pignistic betting distance can be easily obtained using the PPT.

\begin{definition}
Let $m_1$ and $m_2$ be two BPAs on the same frame of discernment $\Theta$ and let $BetP_{m_1}$ and $BetP_{m_2}$ be the results of two pignistic transformations from them respectively. Then the pignistic probability distance between $BetP_{m_1}$ and $BetP_{m_2}$ is defined as
\begin{equation}\label{difBetP}
difBetP = \mathop {\max }\limits_{A \in {2^{\Theta}} } (\left| {BetP_{m_1 } (A) - BetP_{m_2 } (A)} \right|)
\end{equation}

\end{definition}

\subsection{Liu's conflict model}

In \cite{Liu2006Analyzing}, Liu noted that the classical conflict coefficient $k$ cannot effectively measure the degree of conflict between two bodies of evidence. A two-dimensional conflict model is proposed by Liu \cite{Liu2006Analyzing}, in which the pignistic betting distance and the conflict coefficient $k$ are united to represent the degree of conflict.

\begin{definition}
Let $m_1$ and $m_2$ be two BPAs on the same frame of discernment $\Theta$. The two-dimensional conflict model is represented by:
\begin{equation}\label{cf}
cf(m_1 ,m_2 ) = \langle k,difBetP\rangle
\end{equation}
Where $k$ is the classical conflict coefficient of Dempster combination rule in Eq. (\ref{dempster k}), and $difBetP$ is the pignistic betting distance in Eq. (\ref{difBetP}). Iff both $k>\varepsilon$ and $difBetP>\varepsilon$, $m_1$ and $m_2$ are defined as in conflict, where $\varepsilon$ is the threshold of conflict tolerance.
\end{definition}

Liu's conflict model simultaneously considers two parameters to realize conflict management. To some extent, the two-dimensional conflict model could effectively discriminate the degree of conflict. But in most cases, an accurate value, which represents the degree of conflict, is needed for the following process, such as evaluate the reliability of the evidence with assigning different weights.

\subsection{Correlation coefficient of evidence}

In \cite{song2014coefficient}, Song \emph{et al}. proposed a correlation coefficient for the relativity between two BPAs, which can be used to measure the conflict between two BPAs.

\begin{definition}
Let $m_1$ and $m_2$ be two BPAs on the same frame of discernment $\Theta$, containing N mutually exclusive and exhaustive hypotheses. Use the Jaccard matrix $D$, defined in Eq. (\ref{dij}), to modify the BPA:
\begin{equation}\label{BPA_Modify}
\left\{ {\begin{array}{*{20}c}
   {m'_1  = m_1 D}  \\
   {m'_2  = m_2 D}  \\
\end{array}} \right.
\end{equation}
Then the correlation coefficient between two bodies of evidence is defined as:
\begin{equation}\label{Sony_cor}
cor(m_1 ,m_2 ) = \frac{{\langle m'_1 ,m'_2 \rangle }}{{\left\| {m'_1 } \right\| \cdot \left\| {m'_2 } \right\|}}
\end{equation}
where ${\langle m'_1 ,m'_2 \rangle }$ is the inner product of vectors, ${\left\| {m'_1 } \right\|}$ is the norm of vector.
\end{definition}

Song \emph{et al.}'s correlation coefficient measures the degree of relevance between two bodies of evidence: the higher the conflict is, the lower the value of the correlation coefficient is. In Song \emph{et al.}'s correlation coefficient, the Jaccard matrix $D$ is used to modify BPA in order to process BPA including the multi-element subsets. But this modification will repeatedly allocate the belief value of BPA , so the modified BPA does not satisfy the condition $\sum\limits_{A \in {2^{\Theta}}} {m(A)}  = 1$. Thus, Song \emph{et al.}'s correlation coefficient could not satisfy the property $cor(m_1 ,m_2 ) = 1 \Leftrightarrow m_1 {\rm{ = }}m_2$ and sometimes will yield incorrect results.

\section{A new correlation coefficient} \label{proposed method}

The nature of conflict between two BPAs is there exists the difference between the beliefs of two bodies of evidence on the same focal elements, so the conflict could be quantified by the relevance between two bodies of evidence. If the value of the relevance between two bodies of evidence is higher, the degree of the similarity between two bodies of evidence is higher and the degree of conflict between two bodies of evidence is lower; Conversely, if the value of the relevance between two bodies of evidence is lower, the degree of the similarity between two bodies of evidence is lower and the degree of conflict between two bodies of evidence is higher.

In order to measure the degree of relevance between two bodies of evidence, a new correlation coefficient, which considers both the non-intersection and the difference among the focal elements, is proposed. Firstly, some desirable properties for correlation coefficient are shown as follows.

  \begin{definition}\label{properties of correlation coefficient}
  Assume $m_1,\ m_2$ are two BPAs on the same discernment frame $\Theta$, $r_{BPA}(m_1,m_2)$ is denoted as a correlation coefficient for two BPAs, then
  \begin{enumerate}
    \item $r_{BPA}(m_1,m_2)=r_{BPA}(m_2,m_1)$;
    \item $0 \leq r_{BPA}(m_1,m_2)\leq 1$;
    \item if $m_1=m_2$, $r_{BPA}(m_1,m_2)=1$;
    \item $r_{BPA}(m_1,m_2)=0\Leftrightarrow (\bigcup A_i)\bigcap(\bigcup A_j)=\emptyset$, $A_i,\ A_j$ is the focal element of $m_1,\ m_2$, respectively.
  \end{enumerate}
  \end{definition}

In D-S theory, a new correlation coefficient is defined as follows.

\begin{definition}
For a discernment frame $\Theta$ with $N$ elements, suppose the mass of two pieces of evidence denoted by $m_1$, $m_2$. A correlation coefficient is defined as:
  \begin{equation}\label{correlation_coefficient}
r_{BPA}\left( {{m_1},{m_2}} \right) = \frac{{c\left( {{m_1},{m_2}} \right)}}{{\sqrt {c\left( {{m_1},{m_1}} \right) \cdot c\left( {{m_2},{m_2}} \right)} }}
  \end{equation}
Where $c(m_1,m_2)$ is the degree of correlation denoted as:
  \begin{equation}\label{correlation_degree}
   c({m_1},{m_2}) = \sum\limits_{i = 1}^{{2^N}} {\sum\limits_{j = 1}^{{2^N}} {{m_1}({A_i}){m_2}({A_j})\frac{{\left| {{A_i} \cap {A_j}} \right|}}{{\left| {{A_i} \cup {A_j}} \right|}}} }
  \end{equation}
Where $i,j=1,\dots, 2^N$;$A_i$, $A_j$ is the focal elements of mass, respectively; and $\left|  \cdot  \right|$ is the cardinality of a subset.
  \end{definition}

The correlation coefficient $r_{BPA}\left( {{m_1},{m_2}} \right)$ measures the relevance between $m_1$ and $m_2$. The larger the correlation coefficient, the high the relevance between $m_1$ and $m_2$. $r_{BPA}=0$ corresponds to the absence of relevance between $m_1$ and $m_2$, whereas $r_{BPA}=1$ implies $m_1$ and $m_2$ complete relevant, that is, $m_1$ and $m_2$ are identical.

In the following, the mathematical proofs are given to illustrate that the proposed correlation coefficient satisfies all desirable properties defined in Definition \ref{properties of correlation coefficient}. Before the proofs, a lemma is introduced as follows:

  \begin{lemma}\label{lemma}
    For the vector 2-norm, $\forall$ non-zero vector $\xi= (\xi_1,\xi_2,\ldots,\xi_n)^T,$ $\eta=(\eta_1,\eta_2,\ldots,\eta_n)^T$, the condition for the equality in triangle inequality $||\xi+\eta||_2\leq||\xi||_2+||\eta||_2$ is if and only if $\xi=k\eta$.
  \end{lemma}

  \begin{myproof}
  \begin{align*}
     ||\xi+\eta||_2^2 &\leq(||\xi||_2+||\eta||_2)^2 \\
     (\xi_1+\eta_1)^2+(\xi_2+\eta_2)^2+\ldots+(\xi_n+\eta_n)^2 &\leq (\sqrt{\xi_1^2+\xi_2^2+\ldots+\xi_n^2}+\sqrt{\eta_1^2+\eta_2^2+\ldots+\eta_n^2})^2\\
     \xi_1\eta_1+\xi_2\eta_2+\ldots+\xi_n\eta_n &\leq \sqrt{(\xi_1^2+\xi_2^2+\ldots+\xi_n^2)(\eta_1^2+\eta_2^2+\ldots+\eta_n^2)}
    \end{align*}
   From the Cauchy-Buniakowsky-Schwarz Inequality, we can see that the condition of the equality is if and only if $\frac{\xi_1}{\eta_1}=\frac{\xi_2}{\eta_2}=\ldots=\frac{\xi_n}{\eta_n}=k$, namely $\xi=k\eta$.\qed
  \end{myproof}

The proofs are details as follows:

  \begin{myproof}
  Sort the $2^N$ subsets in $\Theta$ as $\{\emptyset,a,b,\ldots,N,ab,ac,\ldots,abc,\ldots\}$, then $m_1,m_2$ are arranged in column vectors in this order,
  \[{{\bf{m}}_1}:x=(m_1(\emptyset),m_1(a),m_1(b),\ldots,m_1(N),m_1(ab),m_1(ac),\ldots)^T,\]
  \[{{\bf{m}}_2}:y=(m_2(\emptyset),m_2(a),m_2(b),\ldots,m_2(N),m_2(ab),m_2(ac),\ldots)^T.\]
  Let $D=\frac{|A_i\cap A_j|}{|A_i\cup A_j|}$, where $A_i,\ A_j\in 2^\Theta$ and the subsets are arranged in the same order as described above. $D$ is positive definite so $\exists C\in R^{2^N\times 2^N}_{2^N}$ and satisfies $D=C^TC$. Then it is obvious that $c(m_1,m_1)=x^TDx=x^TC^TCx$, and similarly $c(m_1,m_2)=x^TC^TCy$, $c(m_2,m_2)=y^TC^TCy$. $r_{BPA}(m_1,m_2)=\frac{x^TC^TCy}{\sqrt{x^TC^TCx}\sqrt{y^TC^TCy}}$.
  \begin{enumerate}
    \item $r_{BPA}(m_1,m_2)=\frac{x^TC^TCy}{\sqrt{x^TC^TCx}\sqrt{y^TC^TCy}}$, $r_{BPA}(m_2,m_1)=\frac{y^TC^TCx}{\sqrt{x^TC^TCx}\sqrt{y^TC^TCy}}$. Because $x^TC^TCy$ is a real number, $x^TC^TCy=(x^TC^TCy)^T=y^TC^TCx$. Thereby, $r_{BPA}(m_1,m_2)=r_{BPA}(m_2,m_1)$.
    \item All the elements in $x,\ y,\ D$ are non-negative real numbers, so it is clear that $r_ {BPA}(m_1,m_2)=\frac{x^TDy}{\sqrt{x^TDx}\sqrt{y^TDy}}\geq 0$.

        Note that $x^TC^TCx=(Cx)^T(Cx)=||Cx||_2^2$, then using the trigonometric inequality on the vector 2-norm for vector $Cx,\ Cy$, the following inequalities are obtained.
        \begin{align*}
         ||C(x+y)||_2^2 &\leq(||Cx||_2+||Cy||_2)^2  \\
         (x+y)^TC^TC(x+y) &\leq(\sqrt{x^TC^TCx}+\sqrt{y^TC^TCy})^2 \\
          x^TC^TCx+x^TC^TCy+y^TC^TCx+y^TC^TCy &\leq x^TC^TCx+y^TC^TCy+2\sqrt{x^TC^TCxy^TC^TCy} \\
          x^TC^TCy &\leq \sqrt{x^TC^TCxy^TC^TCy}\ \ (x^TC^TCy=y^TC^TCx)
        \end{align*}
        Accordingly,
        \begin{equation*}
         r_ {BPA}(m_1,m_2)=\frac{x^TC^TCy}{\sqrt{x^TC^TCx}\sqrt{y^TC^TCy}} \leq 1.
        \end{equation*}
    \item $r_{BPA}(m_1,m_2)=1$, and now $x^TC^TCy=\sqrt{x^TC^TCxy^TC^TCy}$, that is $||C(x+y)|| _2=||Cx||_2+||Cy||_2$. We can get $Cx=kCy$ from Lemma \ref{lemma}. $C$ is an invertible matrix, so $x=ky$. The vectors $x$ and $y$ each represent a BPA, their length are both 1, thus $x=y$, $m_1=m_2$.
    \item If $r_{BPA}(m_1,m_2)=\frac{c(m_1,m_2)}{\sqrt{c(m_1,m_1)\times c(m_2,m_2)}}=0$, then
         \begin{equation*}
        c(m_1,m_2)=\sum_{i=1}^{2^N}\sum_{j=1}^{2^N}m_1(A_i)m_2(A_j)\frac{|A_i\cap A_j|}{|A_i\cup A_j|}=0.
        \end{equation*}
        The above formula shows if $m_1(A_i)m_2(A_j)\neq 0$, namely $A_i,\ A_j$ are the focal elements of $m_1,\ m_2$, respectively, $|A_i\cap A_j|=0$ must occur. In other words, $\forall A_i\cap \forall A_j=\emptyset$ when $A_i,\ A_j$ are the focal elements of $m_1,\ m_2$, respectively. Thereby, $(\bigcup A_i)\bigcap(\bigcup A_j)=\emptyset$ and vice versa.
    \qed
  \end{enumerate}
  \end{myproof}

In summary, the new correlation coefficient satisfies all the desirable properties and could measure the relevance between two bodies of evidence. Based on this, a new conflict coefficient is proposed.

\begin{definition}
  For a discernment frame $\Theta$ with $N$ elements, suppose the mass of two pieces of evidence denoted by $m_1$, $m_2$. A new conflict coefficient between two bodies of evidence ${k_r}$ is defined with the proposed correlation coefficient as:
 \begin{equation}\label{conflict}
\begin{array}{*{20}{l}}
{{k_r}({m_1},{m_2})}& = &{1 - {r_{BPA}}({m_1},{m_2})}\\
{}& = &{1 - \frac{{c({m_1},{m_2})}}{{\sqrt {c({m_1},{m_1}) \cdot c({m_2},{m_2})} }}}
\end{array}
 \end{equation}
\end{definition}

The conflict coefficient ${{k_r}({m_1},{m_2})}$ measures the degree of conflict between $m_1$ and $m_2$. The larger the conflict coefficient, the high the degree of conflict between $m_1$ and $m_2$. $k_r=0$ corresponds to the absence of conflict between $m_1$ and $m_2$, that is, $m_1$ and $m_2$ are identical, whereas $k_r=1$ implies complete contradiction between $m_1$ and $m_2$.

\section{Numerical examples}

In this section, we use some numerical examples to demonstrate the effectiveness of the proposed conflict coefficient.

\textbf{Example 1.} Suppose the discernment frame is $\Theta =\{A_1,A_2,A_3,A_4\}$, two bodies of evidence are defined as following:
\[
\begin{array}{l}
 {\bf{m}}_{\bf{1}} :m_1 (A_1 ,A_2 ) = 0.9,m_1 (A_3 ) = 0.1,m_1 (A_4 ) = 0.0 \\
 {\bf{m}}_2 :m_2 (A_1 ,A_2 ) = 0.0,m_2 (A_3 ) = 0.1,m_2 (A_4 ) = 0.9 \\
 \end{array}
\]
The various conflict measure value are calculated as follows:

The classical conflict coefficient \cite{Dempster1966Upper} $k=0.99$.

Jousselme's evidence distance \cite{jousselme2001new} $d_{BPA}=0.9$.

Song \emph{et al.}'s correlation coefficient \cite{song2014coefficient} $cor=0.3668$.

The proposed correlation coefficient $r_{BPA}=0.0122$.

The proposed conflict coefficient $k_r=1-r_{BPA}=0.9878$.

In this example, the first BPA is almost certain that the true hypothesis is either $A_1$ or $A_2$, whilst the second BPA is almost certain that the true hypothesis is $A_4$. Hence these two BPAs largely contradict with each other, that is, the two BPAs are in high conflict. The results of the classical conflict coefficient $k$, Jousselme's evidence distance $d_{BPA}$ and the proposed conflict coefficient $k_r$ are all consistent with the fact, but Song \emph{et al.}'s correlation coefficient $cor=0.3668$, which means there are a relatively great correlation between these two BPAs. The value of Song \emph{et al}.'s correlation coefficient is unreasonable.

Let us discuss the condition when two BPAs totally contradict. In Example 1, if we revise the two BPAs as:
\[
\begin{array}{l}
 {\bf{m'}}_{\bf{1}} :m'_1 (A_1 ,A_2 ) = 1.0,m'_1 (A_3 ) = 0.0,m'_1 (A_4 ) = 0.0 \\
 {\bf{m'}}_2 :m'_2 (A_1 ,A_2 ) = 0.0,m'_2 (A_3 ) = 0.0,m'_2 (A_4 ) = 1.0 \\
 \end{array}
\]
then these two BPAs totally contradict with each other. This is the situation when the maximal conflict occurs. The results are shown as follows:

The classical conflict coefficient \cite{Dempster1966Upper} $k=1.0$.

Jousselme's evidence distance \cite{jousselme2001new} $d_{BPA}=1.0$.

Song \emph{et al.}'s correlation coefficient \cite{song2014coefficient} $cor=0.3229$.

The proposed correlation coefficient $r_{BPA}=0.0$.

The proposed conflict coefficient $k_r=1-r_{BPA}=1.0$.

In this case, the values of $k$, $d_{BPA}$, and $K_r$ are all equal to 1.0, which indicates that these two BPAs are in total conflict, whilst $cor=0.3229$, which means there are some correlation between the two BPAs. The value of Song \emph{et al.}'s correlation coefficient is unreasonable.

Let us continue to discuss the following two pairs of BPAs.

\textbf{Example 2.} Suppose the discernment frame is $\Theta =\{A_1,A_2,A_3,A_4,A_5,A_6\}$, two pairs of BPAs are defined as following:
\[
\begin{array}{l}
 {\bf{m}}_{\bf{1}} :m_1 (A_1 ) = 0.5,m_1 (A_2 ) = 0.5,m_1 (A_3 ) = 0.0,m_1 (A_4 ) = 0.0 \\
 {\bf{m}}_{\bf{2}} :m_2 (A_1 ) = 0.0,m_2 (A_2 ) = 0.0,m_2 (A_3 ) = 0.5,m_2 (A_4 ) = 0.5 \\
 \end{array}
\]
\[
\begin{array}{l}
 {\bf{m}}_3 :m_3 (A_1 ) = 1/3,m_3 (A_2 ) = 1/3,m_3 (A_3 ) = 1/3,m_3 (A_4 ) = 0.0,m_3 (A_5 ) = 0.0,m_3 (A_6 ) = 0.0 \\
 {\bf{m}}_4 :m_4 (A_1 ) = 0.0,m_4 (A_2 ) = 0.0,m_4 (A_3 ) = 0.0,m_4 (A_4 ) = 1/3,m_4 (A_5 ) = 1/3,m_4 (A_6 ) = 1/3 \\
 \end{array}
\]

The summary of $k$, $d_{BPA}$, $cor$, and $k_r$ values of the two pairs of BPAs is shown in Table \ref{Comparison of Example 2}.

Obviously, these two pairs of BPAs both totally contradict with each other. $k$, and $k_r$ values are equal to 1.0, which is consistent with the fact, whilst $d_{BPA}$, and $cor$ values show that there is some similarity or relevance between the two BPAs, that is, these two BPAs are not in total conflict, especially $cor=0.5606$ for $m_3,m_4$ shows that there is high relevance between $m_3$ and $m_4$. $d_{BPA}$, and $cor$ values are unreasonable. In summary, Jousselme's evidence distance and Song \emph{et al.}'s correlation coefficient could not always give us the correct conflict measurement.

\begin{table}[!ht]
\centering
\caption{Comparison of $k$, $d_{BPA}$, $cor$, and $k_r$ values of the two pairs of BPAs in Example 2}
\label{Comparison of Example 2}
\begin{tabular}{ccccc}
\toprule
  BPAs & $k_r=1-r_{BPA}$ & $k$ & $d_{BPA}$ & $cor$ \\
  \midrule
  ${\bf{m_1}},{\bf{m_2}}$ & 1.0 & 1.0 & 0.7071 & 0.3990 \\
  ${\bf{m_3}},{\bf{m_4}}$ & 1.0 & 1.0 & 0.5774 & 0.5606 \\  \bottomrule
\end{tabular}
\end{table}

On the other hand, the total absence of conflict occurs when two BPAs are identical. In this situation, whatever supported by one BPA is equally supported by the other BPA and there is no slightest difference in their beliefs. The following example is two identical BPAs.

\textbf{Example 3.} Suppose the discernment frame is $\Theta =\{A_1,A_2,A_3,A_4,A_5\}$, two bodies of evidence are defined as following:
\[
\begin{array}{l}
 {\bf{m}}_{\bf{1}} :m_1 (A_1 ) = 0.2,m_1 (A_2 ) = 0.2,m_1 (A_3 ) = 0.2,m_1 (A_4 ) = 0.2,m_1 (A_5 ) = 0.2 \\
 {\bf{m}}_2 :m_2 (A_1 ) = 0.2,m_2 (A_2 ) = 0.2,m_2 (A_3 ) = 0.2,m_2 (A_4 ) = 0.2,m_2 (A_5 ) = 0.2 \\
 \end{array}
\]
In this case, $k_r=0$, $d_{BPA}=0$, and $cor=1$ all indicates that the two BPAs are identical, which are consistent with the fact. But the classical conflict coefficient $k=0.8$, which indicates that these two BPAs are in high conflict. Obviously, the value of $k=0.8$ is incorrect. With regard to this example, using $k$ as a quantitative measure of conflict is not always suitable.

According the above examples, it can be concluded that the proposed correlation coefficient and the proposed conflict coefficient always give the correct quantitative measure of relevance and conflict between two bodies of evidence. In order to further verify the effectiveness of the proposed method, consider the following example:

\textbf{Example 4.} Suppose the number of elements in the discernment frame is 20, such as $\Theta =\{1,2,3,4,...,20\}$, two bodies of evidence are defined as following:

${{\bf{m}}_1}:{m_1}(2,3,4) = 0.05 ,{m_1}(7) = 0.05 ,{m_1}(\Theta) = 0.1, {m_1}({\rm A}) = 0.8$

\begin{figure}[!ht]
\centering
\includegraphics[scale=0.35]{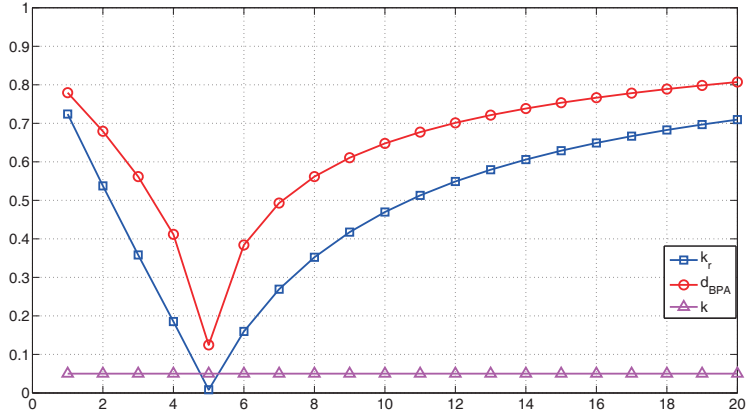}
\caption{Comparison of correlation degree}
\label{Comparison figure}
\end{figure}

${{\bf{m}}_2}:{m_2}(1,2,3,4,5) = 1$
where the \rm A is a variable set taking values as follow:\\
{\rm{\{1\},\{1,2\} \{1,2,3\},\{1,2,3,4\},}}\dots,{\rm{\{1,2,3,4,}}\dots{\rm{,19,20\} }}.
In terms of conflict analysis, the comparative behavior of the two BPAs are shown in Table \ref{Comparison of correlation degree} and Fig. \ref{Comparison figure}. From which we can find the proposed conflict coefficient $k_r$ adopts the similar behavior as Jousselme's evidence distance, that is, when set \rm A tends to the set \{1,2,3,4,5\}, both the values of $k_r$ and $d_{BPA}$ tend to their minimum. On the contrary, the two values will increase when the set \rm A departs from the set \{1,2,3,4,5\}. Fig. \ref{Comparison figure} shows that the trends of $k_r$, and $d_{BPA}$ value are consistent with intuitive analysis, when the size of set \rm A changes, whilst the classical conflict coefficient $k$ fails to differentiate the changes of evidence. So both the proposed conflict coefficient and Jousselme's evidence distance are appropriate to measure the conflict degree of evidence in this example.

However, the major drawback of $d_{BPA}$ is its inability to full consider the non-intersection among the focal elements of BPAs. In Example 2, Jousselme's evidence distance can not give us the correct conflict measurement. As to the classical conflict coefficient $k$, it only takes into consider the non-intersection of the focal elements, but not the difference among the focal elements. Because of the lack of information, $k$ is not sufficient as the quantitative measure of conflict between two BPAs. Compared with Jousselme's evidence distance and the classical conflict coefficient, the proposed correlation coefficient takes into consider both the non-intersection and the difference among the focal elements of BPAs. Therefore, to some extent, the proposed correlation coefficient combine the classical conflict coefficient and Jousselme's evidence distance, and overcome their respective demerits. Thus, the proposed method can measure the degree of relevance and conflict between belief functions correctly and effectively.

\begin{table}[!ht]
\centering
\caption{Comparison of $k_r$, $d_{BPA}$, and $k$ values in Example 4}
\label{Comparison of correlation degree}
\begin{tabular}{lllll}
\toprule
P             & $k_r=1-r_{BPA}$      & $d_{BPA}$    & $k$    &   \\ \hline
\{1\}         & 0.7348 & 0.7858 & 0.05 \\
\{1,2\}       & 0.5483 & 0.6866 & 0.05 \\
\{1,2,3\}     & 0.3690  & 0.5705 & 0.05 \\
\{1,2,3,4\}   & 0.1964 & 0.4237 & 0.05 \\
\{1,2,3,4,5\} & 0.0094 & 0.1323 & 0.05 \\
\{1,2,\dots,6\}    & 0.1639 & 0.3884 & 0.05 \\
\{1,2,\dots,7\}      & 0.2808 & 0.5029 & 0.05 \\
\{1,2,\dots,8\}    & 0.3637 & 0.5705 & 0.05 \\
\{1,2,\dots,9\}    & 0.4288 & 0.6187 & 0.05 \\
\{1,2,\dots,10\}    & 0.4770  & 0.6554 & 0.05 \\
\{1,2,\dots,11\}    & 0.5202 & 0.6844 & 0.05 \\
\{1,2,\dots,12\}    & 0.5565 & 0.7082 & 0.05 \\
\{1,2,\dots,13\}    & 0.5872 & 0.7281 & 0.05 \\
\{1,2,\dots,14\}    & 0.6137 & 0.7451 & 0.05 \\
\{1,2,\dots,15\}    & 0.6367 & 0.7599 & 0.05 \\
\{1,2,\dots,16\}    & 0.6569 & 0.7730  & 0.05 \\
\{1,2,\dots,17\}     & 0.6748 & 0.7846 & 0.05 \\
\{1,2,\dots,18\}    & 0.6907 & 0.7951 & 0.05 \\
\{1,2,\dots,19\}    & 0.7050  & 0.8046 & 0.05 \\
\{1,2,\dots,20\}    & 0.7178 & 0.8133 & 0.05 \\ \bottomrule
\end{tabular}
\end{table}

\section{Conclusions}
In D-S theory, it is necessary to measure the conflicts of belief functions. A correlation coefficient provides a promising way to address the issue. In this paper, a new correlation coefficient of belief functions is presented. It can overcome the drawbacks of the existing methods. Numerical examples in conflicting management are illustrated to show the efficiency of the proposed correlation coefficient of belief functions.

\section*{Acknowledgment}
The work is partially supported by National Natural Science Foundation of China (Grant No. 61671384), Natural Science Basic Research Plan in Shaanxi Province of China (Program No. 2016JM6018), the Fund of Shanghai Aerospace Science and Technology (Program No. SAST2016083).

\section*{References}

\bibliography{myreference}

\begin{thebibliography}{10}
\expandafter\ifx\csname url\endcsname\relax
  \def\url#1{\texttt{#1}}\fi
\expandafter\ifx\csname urlprefix\endcsname\relax\def\urlprefix{URL }\fi
\expandafter\ifx\csname href\endcsname\relax
  \def\href#1#2{#2} \def\path#1{#1}\fi

\bibitem{Dempster1966Upper}
A.~P. Dempster, Upper and lower probabilities induced by a multivalued mapping,
  Annals of Mathematical Statistics 38~(2) (1966) 57--72.

\bibitem{shafer1976mathematical}
G.~Shafer, A mathematical theory of evidence, Vol.~1, Princeton university
  press Princeton, Princeton, NJ, USA, 1976.

\bibitem{Yager2016Evaluating}
R.~R. Yager, N.~Alajlan, {Evaluating Belief Structure Satisfaction to Uncertain
  Target Values}, {IEEE Transactions on Cybernetics} {46}~({4}) ({2016})
  {869--877}.

\bibitem{Majb2016AnEvidential}
J.~Ma, W.~Liu, P.~Miller, H.~Zhou, {An evidential fusion approach for gender
  profiling}, {Information Sciences} {333} ({2016}) {10--20}.

\bibitem{Islam2017Water}
M.~S. Islam, R.~Sadiq, M.~J. Rodriguez, H.~Najjaran, M.~Hoorfar, {Integrated
  Decision Support System for Prognostic and Diagnostic Analyses of Water
  Distribution System Failures}, {Water Resources Management} {30}~({8})
  ({2016}) {2831--2850}.

\bibitem{Moenks2016Information}
U.~Moenks, H.~Doerksen, V.~Lohweg, M.~Huebner, {Information Fusion of
  Conflicting Input Data}, {SENSORS} {16}~({11}).
\newblock \href {http://dx.doi.org/{10.3390/s16111798}}
  {\path{doi:{10.3390/s16111798}}}.

\bibitem{Perez2016Using}
A.~Perez, H.~Tabia, D.~Declercq, A.~Zanotti, {Using the conflict in
  Dempster-Shafer evidence theory as a rejection criterion in classifier output
  combination for 3D human action recognition}, {Image and Vision Computing}
  {55}~({2, SI}) ({2016}) {149--157}.
\newblock \href {http://dx.doi.org/{10.1016/j.imavis.2016.04.010}}
  {\path{doi:{10.1016/j.imavis.2016.04.010}}}.

\bibitem{Denoeux2016Evidential}
T.~Denoeux, S.~Sriboonchitta, O.~Kanjanatarakul, {Evidential clustering of
  large dissimilarity data}, {Knowledge-Based Systems} {106} ({2016})
  {179--195}.
\newblock \href {http://dx.doi.org/{10.1016/j.knosys.2016.05.043}}
  {\path{doi:{10.1016/j.knosys.2016.05.043}}}.

\bibitem{Pedrycz2016Data}
S.~An, Q.~Hu, W.~Pedrycz, P.~Zhu, E.~C.~C. Tsang, {Data-Distribution-Aware
  Fuzzy Rough Set Model and its Application to Robust Classification}, {IEEE
  Transactions on Cybernetics} {46}~({12}) ({2016}) {3073--3085}.

\bibitem{Pedrycz2016Multiobjective}
P.~Ekel, I.~Kokshenev, R.~Parreiras, W.~Pedrycz, J.~Pereira, Jr.,
  {Multiobjective and multiattribute decision making in a fuzzy environment and
  their power engineering applications}, {Information Sciences} {361} ({2016})
  {100--119}.

\bibitem{Frikha2014On}
A.~Frikha, {On the use of a multi-criteria approach for reliability estimation
  in belief function theory}, {Information Fusion} {18} ({2014}) {20--32}.
\newblock \href {http://dx.doi.org/{10.1016/j.inffus.2013.04.010}}
  {\path{doi:{10.1016/j.inffus.2013.04.010}}}.

\bibitem{Zadeh1986A}
L.~A. Zadeh, A simple view of the dempster-shafer theory of evidence and its
  implication for the rule of combination, Ai Magazine 7~(2) (1986) 85--90.

\bibitem{Chin2015Weighted}
K.~S. Chin, C.~Fu, Weighted cautious conjunctive rule for belief functions
  combination, Information Sciences 325 (2015) 70--86.

\bibitem{Yang2013Evidential}
J.~B. Yang, D.~L. Xu, Evidential reasoning rule for evidence combination,
  Artificial Intelligence 205~(12) (2013) 1--29.

\bibitem{DengXY2016IEEE}
X.~Deng, D.~Han, J.~Dezert, Y.~Deng, Y.~Shyr, {Evidence Combination From an
  Evolutionary Game Theory Perspective}, {IEEE Transactions on Cybernetics}
  {46}~({9}) ({2016}) {2070--2082}.

\bibitem{Yang2016Anew}
Y.~Yang, D.~Han, {A new distance-based total uncertainty measure in the theory
  of belief functions}, {Knowledge Based Systems} {94} ({2016}) {114--123}.

\bibitem{Jiang2016Conflicting}
W.~Jiang, M.~Zhuang, X.~Qin, Y.~Tang, Conflicting evidence combination based on
  uncertainty measure and distance of evidence, SpringerPlus 5~(1) (2016)
  1--11.
\newblock \href {http://dx.doi.org/10.1186/s40064-016-2863-4}
  {\path{doi:10.1186/s40064-016-2863-4}}.

\bibitem{Wangjw2016evidence}
J.~Wang, F.~Xiao, X.~Deng, L.~Fei, Y.~Deng, {Weighted Evidence Combination
  Based on Distance of Evidence and Entropy Function}, {International Journal
  of Distributed Sensor Networks} {12}~({7}).
\newblock \href {http://dx.doi.org/{10.1177/155014773218784}}
  {\path{doi:{10.1177/155014773218784}}}.

\bibitem{Zhao2016anovel}
Y.~Zhao, R.~Jia, P.~Shi, {A novel combination method for conflicting evidence
  based on inconsistent measurements}, {Information Sciences} {367} ({2016})
  {125--142}.
\newblock \href {http://dx.doi.org/{10.1016/j.ins.2016.05.039}}
  {\path{doi:{10.1016/j.ins.2016.05.039}}}.

\bibitem{Lin2016Anew}
Y.~Lin, C.~Wang, C.~Ma, Z.~Dou, X.~Ma, {A new combination method for
  multisensor conflict information}, {Journal of Supercomputing} {72}~({7})
  ({2016}) {2874--2890}.
\newblock \href {http://dx.doi.org/{10.1007/s11227-016-1681-3}}
  {\path{doi:{10.1007/s11227-016-1681-3}}}.

\bibitem{Oliveira2016Amulticriteria}
L.~G. de~Oliveira~Silva, A.~T. de~Almeida-Filho, {A multicriteria approach for
  analysis of conflicts in evidence theory}, {Information Sciences} {346}
  ({2016}) {275--285}.
\newblock \href {http://dx.doi.org/{10.1016/j.ins.2016.01.080}}
  {\path{doi:{10.1016/j.ins.2016.01.080}}}.

\bibitem{Lefevre2013How}
E.~Lefevre, Z.~Elouedi, {How to preserve the conflict as an alarm in the
  combination of belief functions?}, {Decision Support Systems} {56} ({2013})
  {326--333}.
\newblock \href {http://dx.doi.org/{10.1016/j.dss.2013.06.012}}
  {\path{doi:{10.1016/j.dss.2013.06.012}}}.

\bibitem{Destercke2013Toward}
S.~Destercke, T.~Burger, {Toward an Axiomatic Definition of Conflict Between
  Belief Functions}, {IEEE Transactions on Cybernetics} {43}~({2}) ({2013})
  {585--596}.
\newblock \href {http://dx.doi.org/{10.1109/TSMCB.2012.2212703}}
  {\path{doi:{10.1109/TSMCB.2012.2212703}}}.

\bibitem{Sarabi2013Information}
A.~Sarabi-Jamab, B.~N. Araabi, T.~Augustin, {Information-based dissimilarity
  assessment in Dempster-Shafer theory}, {Knowledge-Based Systems} {54}~({SI})
  ({2013}) {114--127}.
\newblock \href {http://dx.doi.org/{10.1016/j.knosys.2013.08.030}}
  {\path{doi:{10.1016/j.knosys.2013.08.030}}}.

\bibitem{Yu2015Animproved}
C.~Yu, J.~Yang, D.~Yang, X.~Ma, H.~Min, {An improved conflicting evidence
  combination approach based on a new supporting probability distance}, {Expert
  Systems with Applications} {42}~({12}) ({2015}) {5139--5149}.
\newblock \href {http://dx.doi.org/{10.1016/j.eswa.2015.02.038}}
  {\path{doi:{10.1016/j.eswa.2015.02.038}}}.

\bibitem{jousselme2001new}
A.-L. Jousselme, D.~Grenier, {\'E}.~Boss{\'e}, A new distance between two
  bodies of evidence, Information fusion 2~(2) (2001) 91--101.

\bibitem{Liu2006Analyzing}
W.~Liu, Analyzing the degree of conflict among belief functions, Artificial
  Intelligence 170~(11) (2006) 909 -- 924.

\bibitem{song2014coefficient}
Y.~Song, X.~Wang, L.~Lei, A.~Xue, {Evidence Combination Based on Credibility
  and Separability}, {International Conference on Signal Processing}, {2014},
  pp. {1392--1396}.

\bibitem{Ma2015Combination}
M.~Ma, J.~An, Combination of evidence with different weighting factors: A novel
  probabilistic-based dissimilarity measure approach, Journal of Sensors 2015
  (2015) 509385.

\bibitem{Zhang2016Stumble}
H.~Zhang, Z.~Liu, G.~Chen, Y.~Zhang, {Stumble Mode Identification of Prosthesis
  Based on the Dempster-Shafer Evidential Theory}, {Chinese Control and
  Decision Conference}, {2016}, pp. {2794--2798}.

\bibitem{Zhu2012Information}
H.~Zhu, Z.~Ma, H.~Sun, H.~Wang, {Information Correlation Entropy Based D-S
  Evidence Theory Used in Fault Diagnosis}, in: {2012 International Conference
  on Quality, Reliability, Risk, Maintenance, and Safety Engineering
  (ICQR2MSE)}, {2012}, pp. {336--338}.

\bibitem{Zhang2012Improvement}
L.~Zhang, J.~Yuan, C.~Zhao, {Improvement of Transformer Gas-in-Oil Diagnosis
  Based on Evidence Theory}, in: {2012 Asia-Pacific Power and Energy
  Engineeting Conference (APPEEC)}.

\bibitem{Zhang2016ANP}
X.~Zhang, Y.~Deng, F.~T.~S. Chan, A.~Adamatzky, S.~Mahadevan, {Supplier
  selection based on evidence theory and analytic network process},
  {Proceedings of the Institution of Mechanical Engineers, Part B: Journal of
  Engineering Manufacture} {230}~({3}) ({2016}) {562--573}.

\bibitem{Jiang2016sensor}
W.~Jiang, C.~Xie, M.~Zhuang, Y.~Shou, Y.~Tang, Sensor data fusion with
  z-numbers and its application in fault diagnosis, Sensors 16~(9) (2016) 1509.
\newblock \href {http://dx.doi.org/10.3390/s16091509}
  {\path{doi:10.3390/s16091509}}.

\bibitem{mo2016generalized}
H.~Mo, X.~Lu, Y.~Deng, A generalized evidence distance, Journal of Systems
  Engineering and Electronics 27~(2) (2016) 470--476.

\bibitem{Smets1999TBM}
P.~Smets, R.~Kennes, The transferable belief model, Artificial Intelligence
  66~(2) (1994) 191--234.

\end{thebibliography}

\end{document}